# Language Resources and Technologies for Non-Scheduled and Endangered Indian Languages

**Ritesh Kumar, Bornini Lahiri**

Dr. Bhimrao Ambedkar University, Indian Institute of Technology - Kharagpur

Agra, Kharagpur

E-mail: ritesh78_llh@jnu.ac.in, bornini@hss.iitkgp.ac.in

**Abstract**

In the present paper, we will present a survey of the language resources and technologies available for the non-scheduled and endangered languages of India. While there have been different estimates from different sources about the number of languages in India, it could be assumed that there are more than 1,000 languages currently being spoken in India. However barring some of the 22 languages included in the 8th Schedule of the Indian Constitution (called the scheduled languages), there is hardly any substantial resource or technology available for the rest of the languages. Nonetheless there have been some individual attempts at developing resources and technologies for the different languages across the country. Of late, some financial support has also become available for the endangered languages. In this paper, we give a summary of the resources and technologies for those Indian languages which are not included in the 8th schedule of the Indian Constitution and/or which are endangered.

## 1. Non-scheduled Indian Languages: Background

India is the most linguistically diverse country across the globe. As per the latest census report (Census 2011), there are 1,369 mother tongues currently being spoken in the country. These mother tongues are grouped and divided into 2 categories -

- **Scheduled Languages**: The Constitution of India includes a list of Indian languages in its 8th schedule. These languages are called scheduled languages. Currently there are 22 languages included in the list including Assamese, Bengali, Bodo, Dogri, Gujarati, Hindi, Kannada, Kashmiri, Konkani, Maithili, Malayalam, Manipuri, Marathi, Nepali, Odia, Punjabi, Sanskrit, Santali, Sindhi, Tamil, Telugu and Urdu. The list is modifiable and new languages may be added as per the political decisions taken by the Government. As of now there have been demands for inclusion of 41 more languages to the list. The Government of India is under obligation to take measures for the development of these languages (in terms of  and as such get financial and policy support from the Government of India for their development in various ways including development of language resources and technologies. As is evident, the support from the Government has resulted in substantial development of languages resources and technologies for most of these languages (as described in the white papers included in this proceeding).

- **Non-Scheduled Languages**: As the name itself suggests, all the languages and mother tongues, besides the 22 scheduled languages, are considered non-scheduled languages. Census of India 2011 lists a group of 99 non-scheduled languages. However, all the non-scheduled languages and mother tongues are not listed under this group. A large number of these non-scheduled languages are also listed as mother tongues of scheduled languages. Furthermore languages with less than 10,000 speakers are grouped under a category called 'Others' and their names and number of speakers are not made public – this results in 'Others' category in each group of scheduled and non-scheduled languages. Furthermore this grouping of multiple languages / mother tongues under one name has led to an erroneous impression that these are varieties / dialects of the language groups mentioned in the Census reports. This impression, coupled with an almost complete lack of funding for all of these languages, has resulted in a situation where there have been hardly any language resources and technologies being developed for these languages. So even though none of the Indian languages are resource-rich in comparison to some of the European or other Asian languages like English, German, Chinese, Japanese and other, non-scheduled languages, are extremely less-resourced in comparison to most of the scheduled languages of the country and most of the times they are completely under-resourced.

## 2. Endangered Indian Languages: Background

In addition to these 2 broad classifications of languages based on their inclusion in the 8th schedule of the Indian Constitution, in the last few years, a substantial discourse has developed around the issues of language shift, language maintenance and language endangerment and death. Because of the several socio-political as well as economic reasons including lack of prestige, lack of access to the education in the language, perceived lack of job opportunities and financial securities for the speakers of the languages, besides several other local and regional factors, a large number of speakers of several languages has started shifting to a more powerful and 'useful' language. Most of the communities which are shifting to one of the major, official and more powerful languages are also comparatively smaller communities. Most of the time the population of these communities is less than 10,000 and as such their language is not even listed in the Census reports. And as the speakers shift to different languages, the language spoken by these communities begin to become threatened very quickly.

The Atlas of World's Languages in Danger (Moseley, 2010), published by UNESCO, posits that India has 196 endangered languages, maximum in any single country across the world. The report generated a lot of interest and concern for the fading voices of the nation. However given the fact that a lot of Indian languages continued to survive with extremely less population and hardly any domain of use, besides the home and the local marketplace, the list given by UNESCO as well as their parameters for determining the vitality of any language came under very sharp criticism from several linguists. Nevertheless it was also widely accepted and acknowledged that even though the number of endangered languages in India may not be as high as UNESCO puts, it is still quite large and a large number of Indian languages currently face threat and extinction.

As a result of an urgent need to document and revitalize the endangered languages and also seize the pace of this quick loss of languages, the Government of India started funding research on these endangered languages. Two major multi-institutional, national projects were initiated -

- **Scheme for Preservation and Protection of Endangered Languages (SPPEL)**: This scheme (see SPPEL website) is currently being run by Central Institute of Indian Languages (CIIL), Mysore. It involves more than 40 different universities across the country along with the in-house personnel of the CIIL. Each university as well as the in-house team of CIIL is involved in the exhaustive documentation of the endangered languages of India in the form of audio-visual recordings, digitized and inter-linearly glossed sentences, discourses, folktales, folk-knowledge, etc., a multilingual, multimodal digital dictionary and a grammatical and ethno-linguistic sketch. In the first phase, 117 languages have been selected for documentation and they are in different stages of completion.

- **Schemes of University Grants Commission**: University Grants Commission (UGC) has floated 2 kinds of schemes aimed towards study and research in endangered languages -

  ◦ Centres for Preservation and Promotion of Endangered Languages in Central Universities: These are individual research-based departments which are working on the documentation and description of endangered languages of India. As of now, several such departments / centres have been established in different central universities of India viz Central University of Jharkhand, Central University of Sikkim, Indira Gandhi National Tribal University (IGNTU), Vishwa Bharati, Tezpur University and others

  ◦ Funding Support to State Universities for Study and Research in Indigenous and Endangered Languages of India: This has similar aims as the centres mentioned in the last paragraph but as the name suggests these are given to the state universities and are structures as research project grants and not research-based departments. As of now, several Universities are working under the ambit of this project viz Jadavpur University, Berhampur University, National Law University and others.

As a result of these various kinds of research fundings and schemes, the research on endangered languages has taken off quite well. Since all of these efforts are aimed towards digital documentation and preparing electronic dictionaries and sketch grammars, even though they are not directly related to producing language resources for the development of language technologies, they are actually producing richly annotated resources which are machine-readable and fit for consumption by the language technology community willing to develop different kinds of technologies for these languages.

### 3. Requirement of Language Resources and Technologies

Language resources are required for developing several different kinds of technologies and applications that can process the human language automatically. In a recent survey conducted jointly by KPMG and Google (KPMG Report), it came to the fore that Indian language internet users have grown from 42 million in 2011 to 234 million in 2016. Moreover, this number is higher than the 175 million users who use the internet in English. The report further states that 60% of Indian language internet users shy away from adopting online services because of the lack of content in their language, thereby, stressing the need for the development of better technologies for Indian languages. Along with this, the presence of technologies and the availability of content on the web is expected to have a positive influence on language prestige as well. Thus there is a need for the development of language resources on technologies from both the perspectives –

- The business perspective where the organizations would like to attract the non-English audience to their products and services.

- The linguistic perspective where the users are encouraged to keep using their language and not shift to the other languages (Lahiri 2018).

In order to develop the technologies, the most basic kinds of resources include

- **Text and speech corpora**: These large, systematic, structured collection of high-quality text or speech recordings need to be annotated with different kinds of information such as part-of-speech categories, morphological information, syntactic information (also called tree-banks) and also semantic information (such as different senses of a word) in order to become usable for the development of different kinds of language technologies. The corpora could also be monolingual and parallel.

- **Lexical and lexico-semantic resources**: These include machine-readable and searchable dictionaries, wordnets, framenets, etc. These resources are required for developing certain kinds of applications which either work with words or with different kinds of semantic and lexical relationships across words.

- **Grammars**: These include the computational grammars as well as the descriptive grammars since both of these could prove to be very useful for developing language technologies.

Some of the most basic technologies and applications that might be developed include the following –

- **Technologies for digital content generation**: These are the most basic technologies that are a pre-requisite for the language to be used on digital platforms and on the internet. These include –
    - Language checking tools including spell checkers and grammar checkers
    - Language typing tools such as keyboards and fonts
    - Speech-to-text systems
    - Text-to-speech systems so as to make the digital content accessible to all.

- **Technologies for grammatical analysis**: These are the technologies that could analyze the grammatical and structural properties of a language, thereby, paving the way for the development of more sophisticated applications. Some of these technologies include
    - Part-of-speech taggers
    - Morphological Analysers
    - Syntactic Parsers
    - Named Entity Recognisers
    - Word Sense Disambiguation System
    - Text summarization systems

- **Language technology-based applications**: These are the applications that make use of language technologies for providing different

kinds of services and applications to the users. Some of the major applications include

- Machine Translation
- Sentiment Analysis
- Computer-assisted language learning
- Information retrieval
- Information extraction
- Question answering and dialogue systems
- Electronic Assistants
- Optical Character Recognition (OCR)

In the following 2 sections we will give an overview of the availability of the resources and technologies in the non-scheduled and endangered languages of India. In order to enable cross-linguistic comparison and inspired by the White Papers published by European Union's META-NET project (META-NET White Paper Series), we have given a rating to each of the language resource of each of the significant language on a scale of 0 to 6 where 0 is very low and 6 very high on each of the following seven criteria–

- **Quantity**: This score is based on how much data is available for the given resource in the given language.
- **Availability**: This score tells us if the resource / technology is available for use by other researchers and how easy it is to get those. So a publicly available resource that could be directly downloaded from the web and that has no restrictions on its usage will get the highest score while the one which is not available for other researchers by any means will get the lowest score.
- **Quality**: This score tells us how well-structured, arranged and clean the resource is. If the resource could be used out-of-the-box without any preprocessing then it gets the highest score. If it is almost impossible to use it then it gets the lowest score.
- **Coverage**: It tells us how well the different aspects of the concerned resource represented. For example, if a POS tagger could tag data from any domain almost equally well then it will get a good coverage score.

- **Maturity**: It tells us for how long the resource is being used. The higher the number of users, the higher is its maturity score.
- **Sustainability**: This score tells for how long the resource will continue to be relevant.
- **Adaptability**: It tells how well the resource could adapt and extend itself to new situations and scenarios.

As one would notice, the scores for each of these categories for most of the language resources are towards the lower side. This is a matter of concern. However given the fact that most of the languages did not make it to the list (thereby implying that they do not have any resource at all), inclusion of the language in this list is significant. Furthermore most of the research on non-scheduled and endangered languages of India has started over a period of the last decade or so and as such this development is quite encouraging.

Since some of the above-mentioned technologies are not available for any of the languages, they are not included in our description. Thus for non-scheduled languages we give scores for only 4 kinds of tools and technologies – Machine Translation, Text-to-speech (TTS), OCR and grammatical analysers (these include part-of-speech taggers and morphological analysers). Similarly, for endangered languages we came across only grammatical analyser for one of the languages and so score for only grammatical analyser is reported. The score for all such technologies not included in the tables should be assumed 0 for all the languages.

We also give a score to each of the languages included in the list based on 2 criteria – availability of corpora and availability of machine-readable dictionaries as well as other related resources - using the following six-point scale.

1. Excellent support
2. Good Support
3. Moderate Support
4. Fragmentary Support
5. Weak Support
6. No Support

We have slightly modified the META-NET white papers scale to separate out weak and no support since this is the scale where most of the non-scheduled and endangered Indian languages lie and clubbing these 2 would mean glossing over the important distinction between a completely unresourced and marginally resourced languages. This scale, however, here also shows the quantum of language technology and resource support available for different languages. This is to be noted that languages without any kind of support is excluded from the list.

## 4. Availability of Language Resources for Non-scheduled Languages

In our survey about the non-scheduled languages, we found out that only 10 non-scheduled languages of the country have some kind of language resource or technology available for it. These languages are as below

- **Magahi**: It is an Eastern Indo-Aryan language spoken in Bihar, Jharkhand and West Bengal. The language currently has a part-of-speech annotated text corpus of approximately 70,000 tokens and a total raw corpus of approximately 17 lakh tokens, speech corpus of approximately 15 hours of naturally occurring conversation, a part-of-speech tagger, a morphological analyser a sentence-level language identification system and a multiword expression identification system (Kumar, Lahiri and Alok, 2012, 2014; Kumar, Behera and Jha, 2017) . OCR for the language is available because of the use of Devanagari for writing the language.

- **Bhojpuri**: It is also an Eastern Indo-Aryan language, spoken in Bihar and Eastern Uttar Pradesh. The language currently has a text corpus of 15,000 sentences, a Sanskrit-Bhojpuri parallel corpus of 10,000 tokens, a Hindi-Bhojpuri parallel corpus of 3,000 tokens, a part-of-speech tagger, a sentence-level language identification system, a Sanskrit-Bhojpuri and a Hindi-Bhojpuri machine translation system and an English-Bhojpuri machine translation system is currently under development (Kumar et al., 2018; Singh and Jha, 2015; Sinha, 2017; Mishra, 2015). Again OCR for the language is available because of the use of Devanagari for writing the texts.

- **Awadhi**: Awadhi is a Central Indo-Aryan language, spoken in Eastern Uttar Pradesh. The language currently has a part-of-speech annotated corpus of approximately 10,000 sentences, a part-of-speech tagger and a sentence-level language identification system (Kumar, et al, 2013; Kumar and Basit, 2018). As for Magahi and Bhojpuri, OCR for the language is available since it uses Devanagari as the script.

- **Braj Bhasha**: Braj Bhasha is a Western Indo-Aryan language, spoken in Western Uttar Pradesh. The language currently has a corpus of over 15,000 sentences, a sentence-level language identification system and a part-of-speech tagger and a morphological analyser are currently under preparation (Kumar, et al. 2018). As with the other languages, OCR for the language available since it also uses Devanagari script.

- **Garhwali**: It is a pahari Indo-Aryan language, spoken in Uttarakhand. A English-Garhwali parallel corpus and English – Garhwali machine translation system are currently under preparation and will be made available very soon.

- **Nyishi**: It is a Tibeto-Burman language, spoken in Arunachal Pradesh. A digital, searchable dictionary of over 700 words is currently available online (Saha and Jha, 2007).

- **Garo**: Garo is also a Tibeto-Burman language, spoken in Meghalaya and parts of Assam. Currently a pronunciation lexicon of 15,000 words currently under preparation. Out of these 11,000 words are already done and available online (Redmon and Sangma, 2018).

- **Sambalpuri**: It is an Eastern Indo-Aryan language, spoken in Orissa. The language currently has a POS-tagged corpus of over 1,21,000 tokens and an automatic POS tagger (Behera, Ojha and Jha, 2018).

- **Rajasthani**: It is a Western Indo-Aryan language, spoken in Rajasthan. The language currently has a text-to-speech system.

- **Limbu**: It is a Tibeto-Burman language which has a bilingual digital dictionary of over 2000 words and annotated recordings of the language. Grammar is also available.

The scores of each of the languages are summarized in the Tables 1 - 10. The average score for each of the resources for each language is summarized in Table 11.

|  | Quantity | Availability | Quality | Coverage | Maturity | Sustainability | Adaptability | Sum |
|---|---|---|---|---|---|---|---|---|
| **Language Resources: Resources, Data and Corpora** | | | | | | | | |
| **Text Corpora** | 2 | 3 | 2 | 3 | 2 | 2 | 3 | 17 |
| **Speech Corpora** | 1 | 1 | 1 | 1 | 1 | 1 | 1 | 7 |
| **Parallel Corpora** | 0 | 0 | 0 | 0 | 0 | 0 | 0 | 0 |
| **Lexical Resources and Dictionaries** | 0 | 0 | 0 | 0 | 0 | 0 | 0 | 0 |
| **Grammars** | 2 | 2 | 2 | 1 | 1 | 2 | 1 | 11 |
| **Language Technology: Tools, Technologies and Applications** | | | | | | | | |
| **Text-to-speech** | 0 | 0 | 0 | 0 | 0 | 0 | 0 | 0 |
| **OCR** | 4 | 3 | 4 | 5 | 4 | 5 | 5 | 30 |
| **Machine Translation** | 0 | 0 | 0 | 0 | 0 | 0 | 0 | 0 |
| **Grammatical Analysers** | 2 | 3 | 3 | 2 | 2 | 3 | 3 | 18 |

Table 1: Language Resources and Technologies for Magahi

|  | Quantity | Availability | Quality | Coverage | Maturity | Sustainability | Adaptability | Sum |
|---|---|---|---|---|---|---|---|---|
| **Language Resources: Resources, Data and Corpora** | | | | | | | | |
| **Text Corpora** | 1 | 1 | 2 | 3 | 1 | 2 | 3 | 13 |
| **Speech Corpora** | 0 | 0 | 0 | 0 | 0 | 0 | 0 | 0 |
| **Parallel Corpora** | 2 | 1 | 3 | 3 | 1 | 2 | 3 | 15 |
| **Lexical Resources and Dictionaries** | 0 | 0 | 0 | 0 | 0 | 0 | 0 | 0 |
| **Grammars** | 3 | 2 | 2 | 1 | 1 | 2 | 1 | 12 |
| **Language Technology: Tools, Technologies and Applications** | | | | | | | | |
| **Text-to-speech** | 0 | 0 | 0 | 0 | 0 | 0 | 0 | 0 |
| **OCR** | 4 | 3 | 4 | 5 | 4 | 5 | 5 | 30 |
| **Machine Translation** | 2 | 1 | 2 | 2 | 2 | 2 | 3 | 14 |
| **Grammatical Analysers** | 1 | 1 | 1 | 1 | 1 | 1 | 1 | 7 |

Table 2: Language Resources and Technologies for Bhojpuri

|  | Quantity | Availability | Quality | Coverage | Maturity | Sustainability | Adaptability | Sum |
| --- | --- | --- | --- | --- | --- | --- | --- | --- |
| **Language Resources: Resources, Data and Corpora** | | | | | | | | |
| **Text Corpora** | 1 | 3 | 2 | 1 | 1 | 2 | 3 | 13 |
| **Speech Corpora** | 0 | 0 | 0 | 0 | 0 | 0 | 0 | 0 |
| **Parallel Corpora** | 0 | 0 | 0 | 0 | 0 | 0 | 0 | 0 |
| **Lexical Resources and Dictionaries** | 0 | 0 | 0 | 0 | 0 | 0 | 0 | 0 |
| **Grammars** | 2 | 2 | 1 | 1 | 1 | 1 | 1 | 9 |
| **Language Technology: Tools, Technologies and Applications** | | | | | | | | |
| **Text-to-speech** | 0 | 0 | 0 | 0 | 0 | 0 | 0 | 0 |
| **OCR** | 4 | 3 | 4 | 5 | 4 | 5 | 5 | 30 |
| **Machine Translation** | 0 | 0 | 0 | 0 | 0 | 0 | 0 | 0 |
| **Grammatical Analysers** | 1 | 1 | 1 | 1 | 1 | 1 | 1 | 7 |

Table 3: Language Resources and Technologies for Awadhi

|  | Quantity | Availability | Quality | Coverage | Maturity | Sustainability | Adaptability | Sum |
| --- | --- | --- | --- | --- | --- | --- | --- | --- |
| **Language Resources: Resources, Data and Corpora** | | | | | | | | |
| **Text Corpora** | 1 | 1 | 2 | 1 | 1 | 2 | 3 | 11 |
| **Speech Corpora** | 0 | 0 | 0 | 0 | 0 | 0 | 0 | 0 |
| **Parallel Corpora** | 0 | 0 | 0 | 0 | 0 | 0 | 0 | 0 |
| **Lexical Resources and Dictionaries** | 0 | 0 | 0 | 0 | 0 | 0 | 0 | 0 |
| **Grammars** | 2 | 2 | 1 | 1 | 1 | 1 | 1 | 9 |
| **Language Technology: Tools, Technologies and Applications** | | | | | | | | |
| **Text-to-speech** | 0 | 0 | 0 | 0 | 0 | 0 | 0 | 0 |
| **OCR** | 4 | 3 | 4 | 5 | 4 | 5 | 5 | 30 |
| **Machine Translation** | 0 | 0 | 0 | 0 | 0 | 0 | 0 | 0 |
| **Grammatical Analysers** | 1 | 1 | 1 | 1 | 1 | 1 | 1 | 7 |

Table 4: Language Resources and Technologies for Braj Bhasha

|                                      | Quantity | Availability | Quality | Coverage | Maturity | Sustainability | Adaptability | Sum |
|--------------------------------------|----------|--------------|---------|----------|----------|----------------|--------------|-----|
| **Language Resources: Resources, Data and Corpora** | | | | | | | | |
| **Text Corpora**                     | 0 | 0 | 0 | 0 | 0 | 0 | 0 | 0 |
| **Speech Corpora**                   | 0 | 0 | 0 | 0 | 0 | 0 | 0 | 0 |
| **Parallel Corpora**                 | 1 | 0 | 1 | 1 | 1 | 1 | 1 | 6 |
| **Lexical Resources and Dictionaries** | 0 | 0 | 0 | 0 | 0 | 0 | 0 | 0 |
| **Grammars**                         | 1 | 1 | 1 | 1 | 1 | 1 | 1 | 7 |
| **Language Technology: Tools, Technologies and Applications** | | | | | | | | |
| **Text-to-speech**                   | 0 | 0 | 0 | 0 | 0 | 0 | 0 | 0 |
| **OCR**                              | 4 | 3 | 4 | 5 | 4 | 5 | 5 | 30 |
| **Machine Translation**              | 1 | 0 | 1 | 1 | 1 | 1 | 1 | 6 |
| **Grammatical Analysers**            | 1 | 1 | 1 | 1 | 1 | 1 | 1 | 7 |

Table 5: Language Resources and Technologies for Garhwali

|                                      | Quantity | Availability | Quality | Coverage | Maturity | Sustainability | Adaptability | Sum |
|--------------------------------------|----------|--------------|---------|----------|----------|----------------|--------------|-----|
| **Language Resources: Resources, Data and Corpora** | | | | | | | | |
| **Text Corpora**                     | 0 | 0 | 0 | 0 | 0 | 0 | 0 | 0 |
| **Speech Corpora**                   | 0 | 0 | 0 | 0 | 0 | 0 | 0 | 0 |
| **Parallel Corpora**                 | 0 | 0 | 0 | 0 | 0 | 0 | 0 | 0 |
| **Lexical Resources and Dictionaries** | 1 | 1 | 1 | 1 | 1 | 1 | 1 | 7 |
| **Grammars**                         | 2 | 2 | 1 | 2 | 1 | 1 | 1 | 10 |
| **Language Technology: Tools, Technologies and Applications** | | | | | | | | |
| **Text-to-speech**                   | 0 | 0 | 0 | 0 | 0 | 0 | 0 | 0 |
| **OCR**                              | 0 | 0 | 0 | 0 | 0 | 0 | 0 | 0 |
| **Machine Translation**              | 0 | 0 | 0 | 0 | 0 | 0 | 0 | 0 |
| **Grammatical Analysers**            | 0 | 0 | 0 | 0 | 0 | 0 | 0 | 0 |

Table 6: Language Resources and Technologies for Nyishi

|  | Quantity | Availability | Quality | Coverage | Maturity | Sustainability | Adaptability | Sum |
|---|---|---|---|---|---|---|---|---|
| **Language Resources: Resources, Data and Corpora** | | | | | | | | |
| **Text Corpora** | 0 | 0 | 0 | 0 | 0 | 0 | 0 | 0 |
| **Speech Corpora** | 0 | 0 | 0 | 0 | 0 | 0 | 0 | 0 |
| **Parallel Corpora** | 0 | 0 | 0 | 0 | 0 | 0 | 0 | 0 |
| **Lexical Resources and Dictionaries** | 3 | 6 | 3 | 2 | 2 | 3 | 3 | 22 |
| **Grammars** | 2 | 2 | 1 | 2 | 1 | 1 | 1 | 10 |
| **Language Technology: Tools, Technologies and Applications** | | | | | | | | |
| **Text-to-speech** | 0 | 0 | 0 | 0 | 0 | 0 | 0 | 0 |
| **OCR** | 0 | 0 | 0 | 0 | 0 | 0 | 0 | 0 |
| **Machine Translation** | 0 | 0 | 0 | 0 | 0 | 0 | 0 | 0 |
| **Grammatical Analysers** | 0 | 0 | 0 | 0 | 0 | 0 | 0 | 0 |

Table 7: Language Resources and Technologies for Garo

|  | Quantity | Availability | Quality | Coverage | Maturity | Sustainability | Adaptability | Sum |
|---|---|---|---|---|---|---|---|---|
| **Language Resources: Resources, Data and Corpora** | | | | | | | | |
| **Text Corpora** | 1 | 1 | 1 | 1 | 1 | 2 | 1 | 8 |
| **Speech Corpora** | 0 | 0 | 0 | 0 | 0 | 0 | 0 | 0 |
| **Parallel Corpora** | 0 | 0 | 0 | 0 | 0 | 0 | 0 | 0 |
| **Lexical Resources and Dictionaries** | 0 | 0 | 0 | 0 | 0 | 0 | 0 | 0 |
| **Grammars** | 0 | 0 | 0 | 0 | 0 | 0 | 0 | 0 |
| **Language Technology: Tools, Technologies and Applications** | | | | | | | | |
| **Text-to-speech** | 0 | 0 | 0 | 0 | 0 | 0 | 0 | 0 |
| **OCR** | 0 | 0 | 0 | 0 | 0 | 0 | 0 | 0 |
| **Machine Translation** | 0 | 0 | 0 | 0 | 0 | 0 | 0 | 0 |
| **Grammatical Analysers** | 1 | 1 | 1 | 1 | 1 | 1 | 1 | 7 |

Table 8: Language Resources and Technologies for Sambalpuri

|  | Quantity | Availability | Quality | Coverage | Maturity | Sustainability | Adaptability | Sum |
|---|---|---|---|---|---|---|---|---|
| **Language Resources: Resources, Data and Corpora** | | | | | | | | |
| **Text Corpora** | 0 | 0 | 0 | 0 | 0 | 0 | 0 | 0 |
| **Speech Corpora** | 1 | 1 | 1 | 1 | 1 | 1 | 1 | 7 |
| **Parallel Corpora** | 0 | 0 | 0 | 0 | 0 | 0 | 0 | 0 |
| **Lexical Resources and Dictionaries** | 0 | 0 | 0 | 0 | 0 | 0 | 0 | 0 |
| **Grammars** | 0 | 0 | 0 | 0 | 0 | 0 | 0 | 0 |
| **Language Technology: Tools, Technologies and Applications** | | | | | | | | |
| **Text-to-speech** | 1 | 1 | 1 | 1 | 1 | 1 | 1 | 7 |
| **OCR** | 0 | 0 | 0 | 0 | 0 | 0 | 0 | 0 |
| **Machine Translation** | 0 | 0 | 0 | 0 | 0 | 0 | 0 | 0 |
| **Grammatical Analysers** | 0 | 0 | 0 | 0 | 0 | 0 | 0 | 0 |

Table 9: Language Resources and Technologies for Rajasthani

|  | Quantity | Availability | Quality | Coverage | Maturity | Sustainability | Adaptability | Sum |
|---|---|---|---|---|---|---|---|---|
| **Language Resources: Resources, Data and Corpora** | | | | | | | | |
| **Text Corpora** | 0 | 0 | 0 | 0 | 0 | 0 | 0 | 0 |
| **Speech Corpora** | 3 | 2 | 2 | 1 | 2 | 2 | 1 | 13 |
| **Parallel Corpora** | 0 | 0 | 0 | 0 | 0 | 0 | 0 | 0 |
| **Lexical Resources and Dictionaries** | 1 | 3 | 3 | 3 | 2 | 3 | 2 | 17 |
| **Grammars** | 1 | 2 | 4 | 4 | 4 | 4 | 3 | 22 |
| **Language Technology: Tools, Technologies and Applications** | | | | | | | | |
| **Text-to-speech** | 0 | 0 | 0 | 0 | 0 | 0 | 0 | 0 |
| **OCR** | 0 | 0 | 0 | 0 | 0 | 0 | 0 | 0 |
| **Machine Translation** | 0 | 0 | 0 | 0 | 0 | 0 | 0 | 0 |
| **Grammatical Analysers** | 0 | 0 | 0 | 0 | 0 | 0 | 0 | 0 |

Table 10: Language Resources and Technologies for Limbu

|  | Magahi | Bhojpuri | Awadhi | Braj | Garhwali | Rasjasthani | Sambalpuri | Nyishi | Garo | Limbu |
|---|---|---|---|---|---|---|---|---|---|---|
| **Language Resources: Resources, Data and Corpora** | | | | | | | | | | |
| Text Corpora | 17 | 13 | 13 | 11 | 0 | 0 | 8 | 0 | 0 | 0 |
| Speech Corpora | 7 | 0 | 0 | 0 | 0 | 7 | 0 | 0 | 0 | 13 |
| Parallel Corpora | 0 | 15 | 0 | 0 | 6 | 0 | 0 | 0 | 0 | 0 |
| Lexical Resources and Dictionaries | 0 | 0 | 0 | 0 | 0 | 0 | 0 | 7 | 22 | 17 |
| Grammars | 11 | 12 | 9 | 9 | 7 | 0 | 0 | 10 | 10 | 22 |
| **Language Technology: Tools, Technologies and Applications** | | | | | | | | | | |
| Text-to-speech | 0 | 0 | 0 | 0 | 0 | 7 | 0 | 0 | 0 | 0 |
| OCR | 30 | 30 | 30 | 30 | 30 | 0 | 0 | 0 | 0 | 0 |
| Machine Translation | 0 | 14 | 0 | 0 | 6 | 0 | 0 | 0 | 0 | 0 |
| Grammatical Analysers | 18 | 7 | 7 | 7 | 7 | 0 | 7 | 0 | 0 | 0 |

Table 11: Summary of language resources and technologies for non-scheduled Indian languages

## 5. Availability of Language Resources for Endangered Languages

In our survey about the endangered Indian languages, we found out that more than 50 endangered languages of the country either already have some kind of language resource or it is under development. Some of these languages are – Great Andamanese, Toto, Beda, Dirang Monpa, Gutub Gadba, Dhimal, Koda and others. However, barring a verb analyser for Great Andamanese, none of these languages have any kind of language technology made available for them. In the tables 12 – 18, we give a summary of the resources available for these representative languages. These languages are as below –

- **Great Andamanese**: It is an endangered language of the Andaman Islands. There is a multilingual digital dictionary with sounds, a verb analyser, grammar and documented data available in the language (Chaudhary, 2008).

- **Toto**: It is a Tibeto-Burman language, spoken in a small village called Totopara in West Bengal. Two multilingual digital dictionaries with around 1500 words have been created by two different agencies. There is a very small parallel corpus of 1000 sentences available in this language. Some phonological descriptions and some speech data are available.

- **Beda**: It is a Tibeto-Burman language. Speech data of 20 hours and a digital dictionary of around 800 words is available.

- **Dirang Monpa**: It is spoken in three small districts of Arunachal Pradesh. It is a Tibeto-Burman language. It has a digital dictionary under process which has around 1000 words at present and some grammatical descriptions along with some speech data of 60 hours and a list of 500 words.

- **Gutub Gadba**: It is an Austro-Asiatic language. It is spoken in parts of Andhra

Pradesh and Orissa. It has language data (words and sentences) available in digital form along with a glossary of words of around 500 words, created by two different sources.

- **Dhimal**: It is a Tibeto-Burman language spoken in Darjeeling district of West Bengal. It has a grammar and a bilingual non-digital dictionary of 2000 words. Other than that it has speech data of 80 hours.

- **Koda**: It is an Austro-Asiatic language spoken in pockets of West Bengal. It has a digital multilingual dictionary of 1500 words. There is documented data available in the language.

Since the state of language resources for all these languages are almost equivalent and it is not feasible to include a separate table for each of these languages in this paper, these could be taken as a good proxy for the state of the other endangered languages as well. While reading the tables, it must be noted that a collection of interlinear glossed sentences is taken as a corpus annotated with part-of-speech and morphological information. Similarly, free translations of the sentences in English create a Language-English parallel corpus. Also the recordings are taken as the speech corpus. Most of this data is digitized in a structured, XML-based format and so these resources are of high-quality and may be used for the development of language technologies.

|  | Quantity | Availability | Quality | Coverage | Maturity | Sustainability | Adaptability | Sum |
|---|---|---|---|---|---|---|---|---|
| **Language Resources: Resources, Data and Corpora** | | | | | | | | |
| **Text Corpora** | 5 | 4 | 5 | 5 | 5 | 5 | 4 | 33 |
| **Speech Corpora** | 4 | 3 | 4 | 4 | 4 | 4 | 4 | 27 |
| **Parallel Corpora** | 3 | 3 | 5 | 5 | 5 | 5 | 5 | 31 |
| **Lexical Resources and Dictionaries** | 3 | 5 | 5 | 5 | 5 | 5 | 5 | 32 |
| **Grammars** | 2 | 4 | 4 | 4 | 4 | 4 | 4 | 26 |
| **Language Technology: Tools, Technologies and Applications** | | | | | | | | |
| **Grammatical Analysers** | 1 | 3 | 3 | 1 | 2 | 2 | 2 | 14 |

Table 12: Language Resources and Technologies for Great Andamanese

|  | Quantity | Availability | Quality | Coverage | Maturity | Sustainability | Adaptability | Sum |
|---|---|---|---|---|---|---|---|---|
| **Language Resources: Resources, Data and Corpora** | | | | | | | | |
| **Text Corpora** | 2 | 0 | 3 | 3 | 2 | 3 | 4 | 17 |
| **Speech Corpora** | 2 | 0 | 3 | 3 | 2 | 3 | 4 | 17 |
| **Parallel Corpora** | 2 | 0 | 3 | 3 | 3 | 3 | 5 | 18 |
| **Lexical Resources and Dictionaries** | 2 | 0 | 3 | 3 | 3 | 3 | 4 | 18 |
| **Grammars** | 1 | 3 | 2 | 2 | 2 | 2 | 3 | 15 |

Table 13: Language Resources and Technologies for Toto

|  | Quantity | Availability | Quality | Coverage | Maturity | Sustainability | Adaptability | Sum |
| --- | --- | --- | --- | --- | --- | --- | --- | --- |
| **Language Resources: Resources, Data and Corpora** | | | | | | | | |
| **Text Corpora** | 1 | 1 | 1 | 1 | 1 | 1 | 1 | 7 |
| **Speech Corpora** | 1 | 0 | 4 | 3 | 3 | 3 | 5 | 19 |
| **Parallel Corpora** | 1 | 1 | 1 | 1 | 1 | 1 | 1 | 7 |
| **Lexical Resources and Dictionaries** | 1 | 1 | 2 | 2 | 2 | 2 | 4 | 14 |
| **Grammars** | 0 | 0 | 0 | 0 | 0 | 0 | 0 | 0 |

Table 14: Language Resources and Technologies for Beda

|  | Quantity | Availability | Quality | Coverage | Maturity | Sustainability | Adaptability | Sum |
| --- | --- | --- | --- | --- | --- | --- | --- | --- |
| **Language Resources: Resources, Data and Corpora** | | | | | | | | |
| **Text Corpora** | 2 | 2 | 4 | 2 | 3 | 4 | 4 | 21 |
| **Speech Corpora** | 2 | 2 | 4 | 2 | 3 | 4 | 4 | 21 |
| **Parallel Corpora** | 2 | 2 | 3 | 2 | 2 | 2 | 4 | 20 |
| **Lexical Resources and Dictionaries** | 3 | 2 | 3 | 3 | 3 | 4 | 4 | 22 |
| **Grammars** | 0 | 0 | 0 | 0 | 0 | 0 | 0 | 0 |

Table 15: Language Resources and Technologies for Dirang Monpa

|  | Quantity | Availability | Quality | Coverage | Maturity | Sustainability | Adaptability | Sum |
| --- | --- | --- | --- | --- | --- | --- | --- | --- |
| **Language Resources: Resources, Data and Corpora** | | | | | | | | |
| **Text Corpora** | 1 | 2 | 4 | 3 | 3 | 3 | 4 | 20 |
| **Speech Corpora** | 1 | 2 | 4 | 3 | 3 | 3 | 4 | 20 |
| **Parallel Corpora** | 1 | 2 | 4 | 3 | 3 | 3 | 4 | 20 |
| **Lexical Resources and Dictionaries** | 1 | 0 | 3 | 3 | 3 | 3 | 4 | 17 |
| **Grammars** | 0 | 0 | 0 | 0 | 0 | 0 | 0 | 0 |

Table 16: Language Resources and Technologies for Gutub Gadba

|  | Quantity | Availability | Quality | Coverage | Maturity | Sustainability | Adaptability | Sum |
|---|---|---|---|---|---|---|---|---|
| **Language Resources: Resources, Data and Corpora** | | | | | | | | |
| **Text Corpora** | 2 | 2 | 4 | 3 | 4 | 3 | 5 | 23 |
| **Speech Corpora** | 2 | 2 | 4 | 3 | 4 | 3 | 5 | 23 |
| **Parallel Corpora** | 2 | 2 | 4 | 3 | 4 | 3 | 5 | 23 |
| **Lexical Resources and Dictionaries** | 1 | 0 | 4 | 3 | 4 | 3 | 5 | 26 |
| **Grammars** | 1 | 3 | 4 | 3 | 4 | 3 | 3 | 21 |

Table 17: Language Resources and Technologies for Dhimal

|  | Quantity | Availability | Quality | Coverage | Maturity | Sustainability | Adaptability | Sum |
|---|---|---|---|---|---|---|---|---|
| **Language Resources: Resources, Data and Corpora** | | | | | | | | |
| **Text Corpora** | 2 | 2 | 4 | 5 | 5 | 5 | 5 | 28 |
| **Speech Corpora** | 2 | 2 | 4 | 5 | 5 | 5 | 5 | 28 |
| **Parallel Corpora** | 1 | 0 | 3 | 3 | 3 | 3 | 5 | 18 |
| **Lexical Resources and Dictionaries** | 1 | 0 | 3 | 3 | 3 | 3 | 5 | 18 |
| **Grammars** | 1 | 0 | 3 | 3 | 3 | 3 | 5 | 18 |

Table 18: Language Resources and Technologies for Koda

|  | Great Andamanese | Toto | Beda | Dirang Monpa | Gutub Gadba | Dhimal | Koda |
|---|---|---|---|---|---|---|---|
| **Language Resources: Resources, Data and Corpora** | | | | | | | |
| **Text Corpora** | 33 | 17 | 7 | 21 | 20 | 23 | 28 |
| **Speech Corpora** | 27 | 17 | 19 | 21 | 20 | 23 | 28 |
| **Parallel Corpora** | 31 | 18 | 7 | 20 | 20 | 23 | 18 |
| **Lexical Resources and Dictionaries** | 32 | 18 | 14 | 22 | 17 | 26 | 18 |
| **Grammars** | 26 | 15 | 0 | 0 | 0 | 21 | 18 |

| Language Technology: Tools, Technologies and Applications |||||||| 
|---|---|---|---|---|---|---|---|
| Grammatical Analysers | 14 | 0 | 0 | 0 | 0 | 0 | 0 |

Table 19: Summary of language resources and technologies for non-scheduled Indian languages

## 6. Summing Up

In this paper, we have presented a survey of the language resources and technologies available for the non-scheduled and endangered Indian languages. As we noted earlier, these languages have only negligible resources in comparison to the other resource-rich languages or even the 22 scheduled Indian languages, which are considered less-resourced languages. As such the overall picture that we got is on expected lines. There are 10 scheduled language for which we have some resources and technologies available. Among these, Magahi and Bhojpuri seems to have the best support with multiple resources and technologies currently available while the other languages have very limited support. Barring Rajasthani, Garo and Nyishi, all other languages have some kind of text corpora available. Since most of the languages discussed here are written in Devanagari, they have an advantage in terms of OCR system being available (which is originally developed for processing Hindi). Most of these languages also have some grammatical analysers available for them. However all the languages lack a basic processing tool like spell checker and an effort in this direction might be useful and helpful. One of the major issues that came up during the course of this survey is the non-availability of the resources. Most of the resources are not available and, at best, they are very tricky to acquire for one's own research, if not impossible. This unwillingness of researchers to talk about their resources and share those is perplexing since it is pretty obvious that in cases like this where external funding is very limited or completely absent, collaboration among the community is essential to make some significant progress.

The situation of endangered languages is not better than the non-scheduled languages even though these languages now have access to some governmental funding. This is so because of 2 reasons – it is only recently that the research on these languages have started and the research is not oriented towards development of language technologies. The availability of language resources is a major issues for these languages. Since the resources for these languages are created using public money, we expect these to be available in public domain. However, because of certain bureaucratic norms as well as community-related restrictions, it has not become possible. Aside from this, it must also be kept in mind that the requirements of endangered languages are very different from those of non-scheduled languages and as such different kinds of technologies need to be given priority. For example, spell checkers or OCR may not be relevant or useful tools for the endangered languages (since they are generally only spoken languages) but a dictionary or a tool for computer-assisted language learning might prove to be quite useful resources for the community as well as researchers trying to revitalise the language.

## 7. Bibliographical References